\documentclass[letterpaper, 10 pt, conference, 
]{ieeeconf}

\IEEEoverridecommandlockouts                              
\usepackage{makecell}  %
\usepackage{graphics} 
\usepackage{epsfig} 
\usepackage{times} 
\usepackage{amsmath} 
\usepackage{amssymb}  
\usepackage{booktabs,ragged2e}
\usepackage{multirow}
\usepackage[flushleft]{threeparttable}
\usepackage{xcolor}
\usepackage[english]{babel}
\usepackage{lipsum}
\usepackage{hyperref}
\usepackage{verbatim}
\usepackage{array}
\usepackage{tabularx}

\usepackage{color}

\newcommand{\col}{\color{black}}

\newcolumntype{Y}{>{\centering\arraybackslash}X}
\newcolumntype{M}[1]{>{\centering\arraybackslash}m{#1}}

\usepackage[ruled,linesnumbered]{algorithm2e}
\SetKwInput{KwRequire}{Require}

\def\figref#1{Fig.~\ref{#1}}

\title{\LARGE \bf
	Optimizing Indoor Farm Monitoring Efficiency Using UAV: Yield Estimation in a GNSS-Denied Cherry Tomato Greenhouse
}

\author{
	Taewook Park, Jinwoo Lee, Hyondong Oh, \textit{Senior Member, IEEE},\\
	Won-Jae Yun, Kyu-Wha Lee
	\thanks{This work was supported by the Basic Science Research Program through the National Research Foundation of Korea (NRF) funded by the Ministry of Education (2020R1A6A1A03040570), the Korea Institute of Planning and Evaluation for Technology in Food, Agriculture, and Forestry (IPET) through the Smart Farm Innovation Technology Development Program, funded by the Ministry of Agriculture, Food and Rural Affairs (MAFRA) (RS-2025-02219411), and the Technology development Program(RS-2024-00441677) funded by the Ministry of SMEs and Startups(MSS, Korea). \textit{(Corresponding author: Hyondong Oh.)}}
	\thanks{Taewook Park, Jinwoo Lee, and Hyondong Oh are with the Department of Mechanical Engineering, Ulsan National Institute of Science and Technology (UNIST), Ulsan 44919, South Korea (e-mail: wook@unist.ac.kr, jinwoolee2021@unist.ac.kr, h.oh@unist.ac.kr).}
	\thanks{Won-Jae Yun and Kyu-Wha Lee are with the Agricultural Robotics and Artificial Intelligence, Metafarmers, Seoul 08793, South Korea (e-mail: wonjae.yun@metafarmers.ai, kyuwha.lee@metafarmers.ai).}%
}

\begin{document}
    \maketitle
	\thispagestyle{empty}
	\pagestyle{empty}	

	\begin{abstract}
As the agricultural workforce declines and labor costs rise, robotic yield estimation has become increasingly important. While unmanned ground vehicles (UGVs) are commonly used for indoor farm monitoring, their deployment in greenhouses is often constrained by infrastructure limitations, sensor placement challenges, and operational inefficiencies. To address these issues, we develop a lightweight unmanned aerial vehicle (UAV) equipped with an RGB-D camera, a 3D LiDAR, and an IMU sensor. The UAV employs a LiDAR-inertial odometry algorithm for precise navigation in GNSS-denied environments and utilizes a 3D multi-object tracking algorithm to estimate the count and weight of cherry tomatoes. We evaluate the system using two dataset: one from a harvesting row and another from a growing row. In the harvesting-row dataset, the proposed system achieves 94.4\% counting accuracy and 87.5\% weight estimation accuracy within a 13.2-meter flight completed in 10.5 seconds. For the growing-row dataset, which consists of occluded unripened fruits, we qualitatively analyze tracking performance and highlight future research directions for improving perception in greenhouse with strong occlusions.
Our findings demonstrate the potential of UAVs for efficient robotic yield estimation in commercial greenhouses.
			
\vspace{5mm}
	\end{abstract}
		
	
	
\section{Introduction}
Indoor farm monitoring provides crucial data for farm management, supporting decisions related to nutrient adjustments, plant pruning, and labor requirements. However, manual counting and weighing of ripened fruits before harvesting is labor-intensive and may damage the target plants. To obtain statistics more efficiently and non-destructively, recent studies have introduced robots with imaging sensors, coupled with deep learning-based approaches for fruit counting and weight estimation \cite{he2022fruit, van2020crop}.

\begin{figure}[t] 
	\centering
	\vspace{2mm}
	\includegraphics[width=1.0\linewidth]{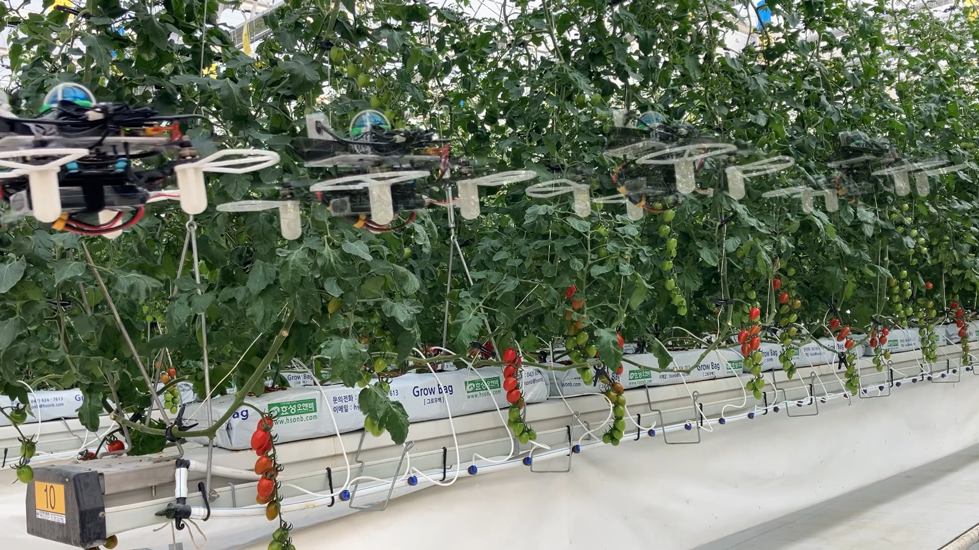}
	\vspace{-6mm}
	\caption{UAV-based monitoring of GNSS-denied greenhouse environment. The proposed system, designed for yield estimation, achieves the accuracy of 94.4\% for fruit counting and 87.5\% for weight estimation, traveling 13.2 meters in just 10.5 seconds. Actual yield estimation computations are performed offline.}
	\label{fig:fig1}
	\vspace{-2mm}
\end{figure}

Unmanned ground vehicles (UGVs) are widely used platforms capable of navigating greenhouse environments to collect dataset for indoor monitoring. These robots maneuver into positions close to target fruits, after which a mounted manipulator adjusts its orientation to achieve an optimal viewpoint, ensuring accurate fruit detection and tracking. Additionally, lifts are frequently employed to control the vertical positioning of the manipulator base \cite{xiong}. 

However, deploying these robots often requires modifications to greenhouse infrastructure. As mobile robots for indoor farms carry manipulators, lifts, batteries, and onboard computers, many deployed UGVs narrowly fit within the constraints of real-world greenhouse applications. For example, hot water pipes are used as guiding tracks to minimize navigation errors. Additional adaptations to the robot's wheels are required as greenhouse floors vary from flat concrete to rough terrain.

Recent advances in camera and LiDAR-based positioning systems, integrated with an IMU, enable precise flight of indoor unmanned aerial vehicles (UAVs). Research on UAV-based navigation in confined spaces \cite{uav_indoor1, uav_indoor2} demonstrates strong stability. Furthermore, recent advancements in UAV assembly components and flight control units (FCUs) have led to smaller UAV sizes, reducing the overall volume compared to earlier models.

In this paper, we leverage advancements in the UAV industry and positioning systems to develop a compact UAV optimized for robotic farm monitoring as shown in \figref{fig:fig1}. Our key contributions are summarized as follows:
\begin{enumerate}
	\item 
	\textit{Design of a Compact Quadcopter for Monitoring in Global Navigation Satellite System (GNSS)-Denied Greenhouses.} The integration of an \mbox{RGB-D} camera, a 3D LiDAR, and IMU measurements enables efficient farm monitoring without a GNSS. The resulting quadcopter weighs 1.4 kg and is able to fly more than 1 kilometers;
	\item 
	\textit{Adaptation of 3D Multi-Object Tracking Algorithms for Cherry Tomatoes.} We review robotic yield estimation methods and adapt 3D multi-object tracking algorithms for the localization and sizing of cherry tomatoes. The algorithm tracks observed tomatoes within a global coordinate system and remains effective even in high-density greenhouse environments; and
	\item \textit{Yield Estimation and Validation in a Commercial Cherry Tomato Farm.} To validate the proposed system, we manually control the UAV with LiDAR-inertial odometry to fly within a narrow corridor. During the flight, \mbox{RGB-D} images and UAV positions are collected to process the yield estimation. After the flight, we harvested the cherry tomatoes of the observed lane to get their actual count and weights. The proposed system achieved the accuracy of 94.4\% for counting and 87.5\% for weighting estimation when inspecting 13.2 m lane only with 10.5 seconds.
\end{enumerate}
\section{Related Work}
\subsection{Efficiency of Robotic Farm Monitoring}

An agricultural robotic platform plays a crucial role in farm monitoring by enabling automatic data acquisition in farm environments. The deployment of such platforms is typically categorized into outdoor and indoor farming environments.

The most extensively studied outdoor robotic platforms include UGVs and UAVs. Several studies \cite{kurtser_robot, massah, zhang} have suggested UGVs as suitable platforms for agricultural robotic systems. These systems incorporate machine vision sensors, enabling robotic platforms to perform yield estimation for crops such as kiwifruit and vine. Additionally, with the integration of manipulators, they can handle more complex tasks such as apple harvesting. In addition to UGV-based robotic platforms, UAVs are recognized for their time efficiency in large-scale agricultural fields \cite{walsh}.  UAV-based robotic systems equipped with vision have been introduced as effective solutions for yield estimation in soybean and wheat cultivation \cite{yu, ali}.

Similar to outdoor environments, UGVs are the primary platforms considered for agricultural robotic systems in indoor farms. Indoor UGVs also integrate vision sensors, manipulators or grippers for complex tasks such as sweet pepper and strawberry harvesting \cite{arad, xiong}. However, UAVs are rarely deployed in indoor farms due to several constraints including the unavailability of external positioning data and challenges in ensuring the safety of humans and plants.

While outdoor farms benefit from both UGV- and UAV-based robotic platforms, indoor farms remain limited to UGV-based solutions, which pose significant challenges for monitoring large-scale or high-rise structured farms. To enhance monitoring efficiency in greenhouses, these challenges require further investigation.

\subsection{Viewpoint Selection for Robotic Agriculture}
Vision-based farm monitoring is widely recognized for its effectiveness, as it captures rich color and semantic information \cite{bargoti}. However, occlusions caused by densely packed plants remain one of the biggest challenges in vision-based farm monitoring \cite{rincon}. Several studies have attempted to address this challenge by developing viewpoint selection strategies mainly with manipulators.

For instance, Kurtser et al. \cite{kurtser_vp} proposed a heuristic algorithm that determines if an additional viewpoint is needed based on occlusion level analysis and the expected number of newly visible fruits, optimizing the time efficiency of pepper harvesting. Similarly, Yi et al. \cite{yi} introduced an active vision strategy that dynamically adjusts the robot's viewpoint to maximize the visibility of fruit stems in grape harvesting tasks. Menon et al. \cite{menon} developed a next-best-view (NBV) \cite{nbv} planner based on shape completion, which predicts missing fruit surfaces and utilizes this information to generate optimized viewpoints for improved reconstruction. Additionally, Burusa et al. \cite{burusa} incorporated semantic-aware NBV planning into an automated harvesting and pruning robotic system, where class labels and confidence scores are used to prioritize relevant plant parts during viewpoint selection.
Rehman et al. \cite{rehman} formulated the viewpoint selection task as a classification problem, where a deep neural network predicts if the camera should move left, right, or remain in place based on the current scene, specifically for sweet pepper harvesting.

Overall, existing methods range from heuristic strategies to deep learning-based approaches, aiming to select multiple viewpoints to overcome occlusions when using manipulators. However, very few studies have addressed the problem of efficient farm monitoring using a UAV.
\section{Methods} 
\subsection{Analysis of Indoor Farm Environments}
\label{sec:env_analysis}

\begin{figure*}[t] 
	\centering
	\includegraphics[width=1.0\linewidth]{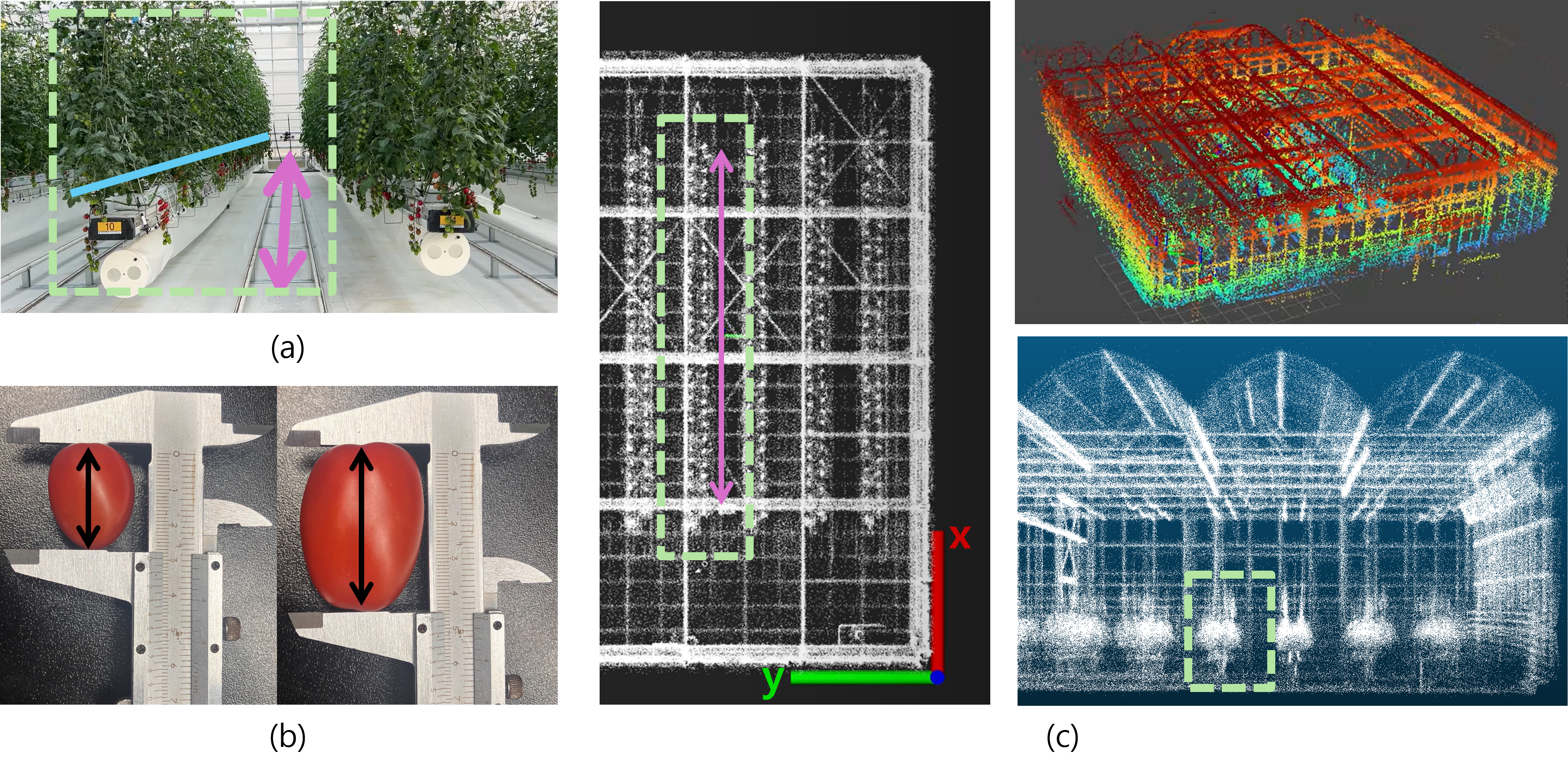}
	\vspace{-6mm}
	\caption{Images to help understanding the indoor farm environment. (a) Validation lane for yield estimation. Blue line indicates a boundary between a harvesting row and a growing row. Pink allows shows the UAV's flight path. All ripened cherry tomatoes in the harvesting row are harvested after the flight. (b) Lengths of the smallest and the biggest cherry tomato among all harvested cherry tomatoes, which are 28mm and 45 mm respectively. (c) Top-view, perspective view, and side-view from the farm's 3D point cloud map. Note that the UAV's flight path is aligned with 3D map's x-axis.}
	\vspace{-4mm}
	\label{fig:farm_analysis}
	\vspace{0mm}
\end{figure*}

\begin{figure}[t] 
	\centering
	\includegraphics[width=1.0\linewidth]{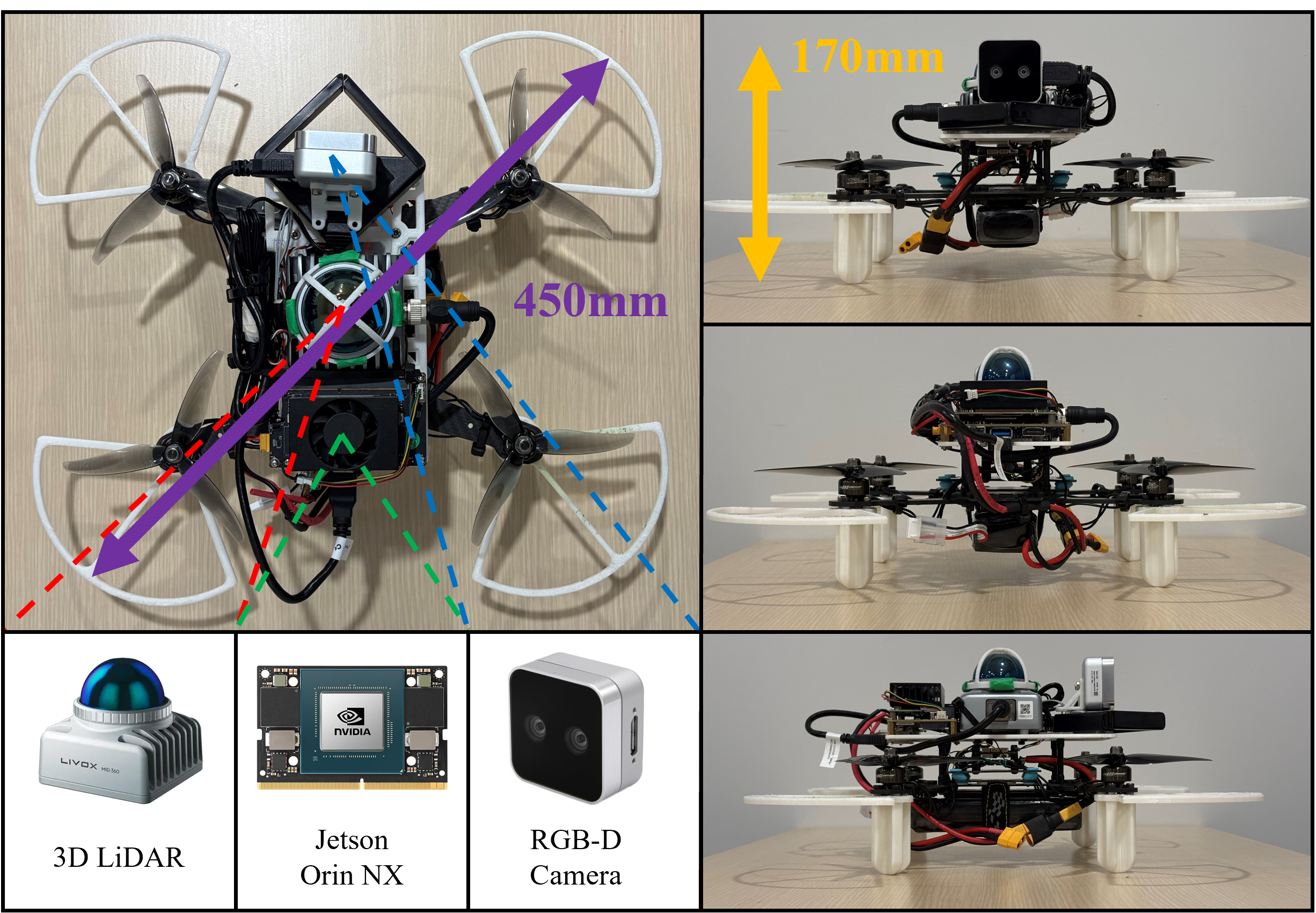}
	\vspace{-6mm}
	\caption{Configurations of the proposed UAV. With a flight speed of 2 m/s and a duration of 9 minutes, the UAV can cover a distance of over 1 km.}
	\vspace{-4mm}
	\label{fig:uav}
	\vspace{0mm}
\end{figure}

To determine key configurations, we first examine the indoor farm environment as illustrated in \figref{fig:farm_analysis}. A 3D point cloud map is utilized to obtain distance and volume measurements. To construct the 3D map, 3D LiDAR scans from the Livox MID-360 and IMU measurements are recorded. We then take advantage of the high robustness of GLIM
\cite{glim}, a LiDAR-inertial odometry estimator that supports 3D mapping.

The total flight path spans 13.2 meter, while the minimum width between two suspended growing troughs is 0.82 meter. Considering that the wheelbase of the proposed UAV is 0.45 meter as shown in \figref{fig:uav}, it must maintain a lateral offset of no more than 0.185 meter from the center to prevent damage to the plants.

\begin{figure*}[tb] 
	\centering
	\includegraphics[width=0.9\linewidth]{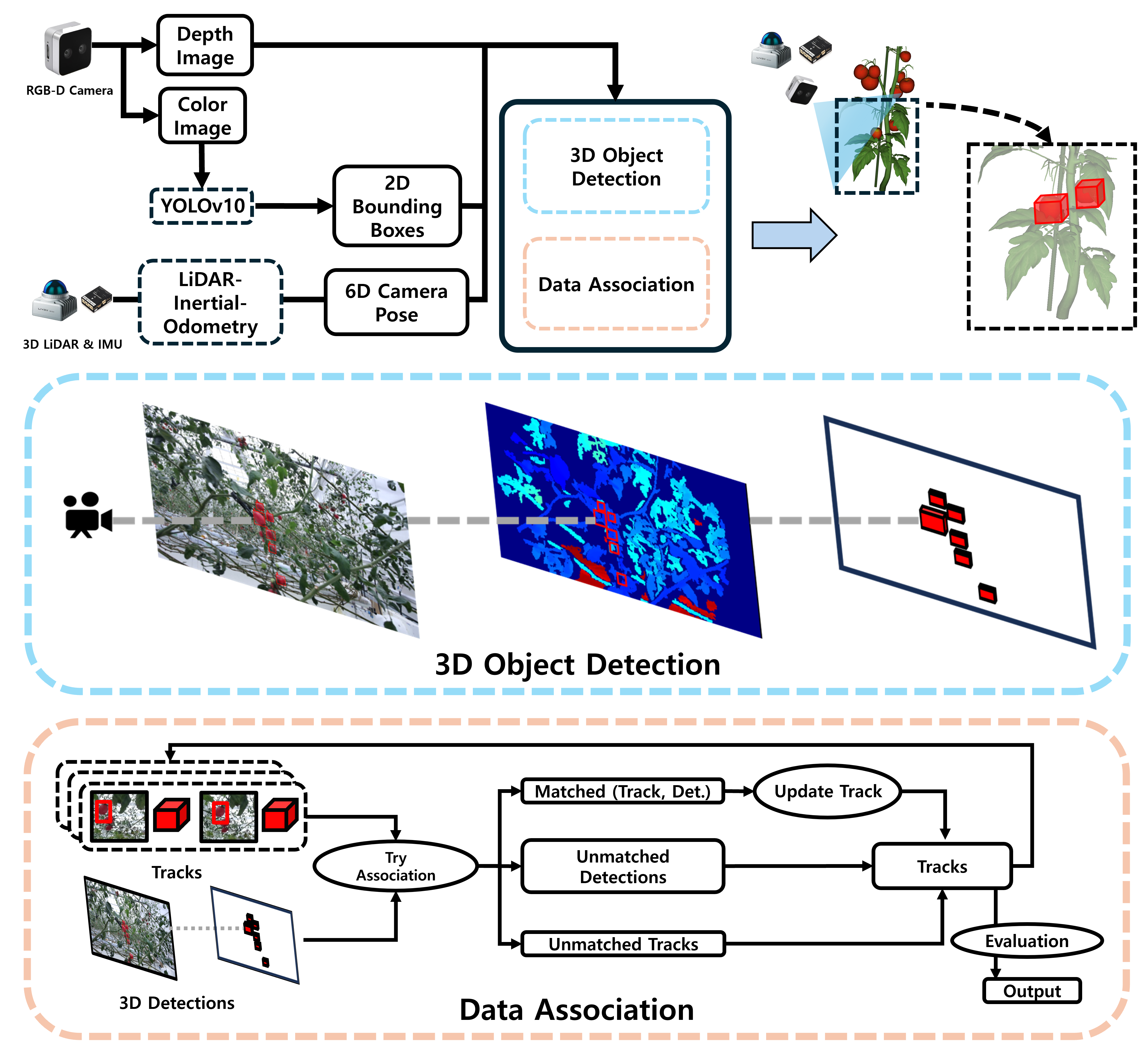}
	\vspace{-5mm}
	\caption{Overview of the proposed 3D multi-object tracking framework. The system takes depth images, 2D bounding boxes from color images, and 6D camera poses as input. By fusing multi-sensor data, the 3D contexts of cherry tomatoes are estimated.  
	\textit{\textbf{Skyblue:}} Creation of 3D detections. Cherry tomatoes are approximated as cubes.  
	\textit{\textbf{Light peach:}} Association rule for tracking. Tracks are created or updated based on the similarity between existing tracks and new 3D detections. After association, the 3D contexts of cherry tomatoes are updated.
	}
	\vspace{-6mm}
	\label{fig:MOT}
\end{figure*}

\subsection{Indoor UAV and Precise Flight}
\label{sec:uav}
The proposed quadrotor platform has a diameter of 22.5 cm including propeller guards, a height of 17 cm, and a net weight of 1.4 kg. It is powered by 3800 kV brushless DC (BLDC) motors and GF 5131 propellers, capable of generating a maximum thrust of 2.4 kg when using a 4S~(14.8~V) Li-Po battery. The MicoAir 743 is used as the flight control unit. With a 3300 mAh Li-Po battery, the platform achieves a maximum flight time of 9 minutes under standard operating conditions. The onboard mission system consists of an NVIDIA Jetson Orin NX computing unit, a Livox MID-360 3D LiDAR, and an Intel RealSense D405 \mbox{RGB-D} camera. IMU measurements are streamed from the flight control unit.

To ensure safe flight without GNSS, external position data must be provided to the FCU. In the proposed UAV, external position estimates are obtained at 10 Hz using a robust LiDAR-inertial odometry estimator, namely GLIM \cite{glim}. Instead of GPU-based computation, which offers higher robustness, we employ a CPU implementation to reduce computational latency. After tuning the covariance matrix of the position estimates required for the FCU to manage the confidence of external position data, robust and stable flight performance is achieved.

\subsection{YOLOv10 and 3D Multi Object Tracking Algorithm}
To estimate cherry tomato yield, we infer the 3D context of each tomato, which consists of a position and a volume. When the UAV records a sensor stream, the same cherry tomato is typically observed multiple times. Therefore, associating multiple observations of the same cherry tomato is necessary. Since the indoor farm contains numerous cherry tomatoes, multiple tomatoes must be identified across different observations. This problem is one of a multi-view perception challenge and can be effectively addressed using a multi-object tracking (MOT) approach \cite{mot1, mot2}.

To successfully track cherry tomatoes, we reviewed existing MOT algorithms in the agricultural domain \cite{mot3, mot4, mot5, mot6, mot7} and traffic surveillance \cite{mot8, mot9, mot10}, identifying common characteristics.  
First, handling noisy detection results plays a crucial role. Deep neural networks applied to color images may fail to detect certain targets and can also produce false alarms.  
Second, 2D MOT, which relies solely on color images and 2D detections, is the most widely used approach \cite{ag_mot_review} due to its relatively simple implementation. However, it struggles when the camera's orientation changes abruptly or when targets become occluded by stems or other objects. Addressing these challenges requires prior knowledge or additional sensor inputs.  
Third, 3D MOT approaches that incorporate not only color images but also depth images, LiDAR scans, or the camera’s 6D poses can address these limitations. However, tracking objects in 3D requires estimating additional states, including their orientations, which significantly increases computational demand.
 
Based on these observations, we decided to track cherry tomatoes in a 3D context by estimating their classes, positions, and volumes. To limit computational complexity and facilitate the easier analysis of algorithm malfunctions, we extensively leverage the stationary nature of plants, as they do not move and therefore do not require dynamic models. Additionally, their orientations are disregarded.

The proposed 3D MOT framework is presented in \figref{fig:MOT}. 2D bounding boxes are detected using a deep neural network. By fusing these 2D bounding boxes with depth images, cherry tomatoes are approximated as cubes. The coordinate system of these cubes is transformed into the global frame using a 6D camera pose. Next, the cubes are associated, and tracks are generated based on similarity measurements. Finally, by evaluating each track, false positive detections are filtered out, and cherry tomatoes are successfully tracked. Details of each processing step are explained in the following paragraphs.

\textit{\textbf{Training YOLOv10.}} To train our network, we collected 2,830 images, with 229,040 instances across five classes labeled by a professional image analysis company. The five classes include tomato stems, unripened tomatoes, ripened tomatoes, leaf branches, and flowers; however, only tomato labels are used in this research. Ripened and unripened tomatoes are labeled only if they exceed 10 pixels in size.

\textit{\textbf{3D Detection of Cherry Tomatoes.}} 2D bounding boxes from an RGB image are projected onto a depth image. Since we used realsense D405, manual calibration of RGB and depth image is not required since it is done internally. The pinhole camera model is then applied to approximate each bounding box as a simple cube. Mathematically, a state \(c\) of a cube is defined as set of a 3D position and its volume.
\setlength{\arraycolsep}{0.0em}
\begin{equation}
	c = \{ X, Y, Z, W, H, L \}.
	\label{eq:cube_def}
\end{equation}

\noindent Here, \( Z \), the distance from the camera to a cherry tomato, is defined as the median depth within the projected ROI.
\setlength{\arraycolsep}{0.0em}
\begin{equation}
	Z = \text{median}(\text{ROI}).
\end{equation}

\noindent Then, other parameters of \(c\) can be calculated as:
\setlength{\arraycolsep}{0.0em}
\begin{gather}
	X_p = \frac{(u_p - c_p) Z}{f_x}, \\
	Y_p = \frac{(v_p - c_p) Z}{f_y}, \\
	W_p = \frac{\Delta u_p \cdot Z}{f_x}, \\
	H_p = \frac{\Delta v_p \cdot Z}{f_y}, \\
	L_p = \frac{W_p + H_p}{2}.
\end{gather}

\noindent Here, \( u_p, v_p, \Delta u_p \), and \( \Delta v_p \) denote the center coordinates, width, and height of a 2D bounding box \(p\) in the image plane, respectively.  
\(f_x, f_y, c_x\), and \(c_y\) represent the focal lengths in the x- and y-directions of the image plane, and the principal point coordinates of the camera.  
\( W_p, H_p, \) and \( L_p \) correspond to the estimated width, height, and length of the 3D object, respectively. {\col Note that the length, defined along the camera's principal axis, is approximated as the average of the width and height to exploit the approximate symmetry of cherry tomatoes.}
Finally, the 6D camera pose is used to transform the cubes' coordinates and volumes into the global frame. It is important to mention that depth values for distant objects or near edges may be invalid. In such cases, the median depth becomes an invalid value, leading to rejection of the 3D detection.

\textit{\textbf{Data Association.}} The proposed 3D object detection module outputs multiple cubes, each characterized by a 3D position, volume, and class ID from the object detector network. However, these detections cannot be directly considered as cherry tomatoes due to the following reasons. First, within a time series of detected cubes, multiple cubes may originate from the same cherry tomato. Second, the YOLOv10 object detector may generate false alarms, producing bounding boxes that do not correspond to real tomatoes. Third, cubes may fail to be detected due to occlusions or unpredictable network inference failures. 

To address these issues, we apply an additional data association process, as illustrated in \figref{fig:MOT}. The proposed method requires previously maintained tracks and newly detected 3D objects. Each track represents a summary of similar cubes assumed to originate from the same tomato. The 3D context of a track is also represented as a cube, with its position and volume computed as a convex combination of the associated cubes. 

The proposed association process proceeds as follows. When new 3D detections with identical timestamps become available, they are associated with existing tracks if the Euclidean distance between them is sufficiently small. Mathematically, a cube \( c_t \) from a track and a cube \( c_d \) from a 3D detection form a matched pair when the following condition is satisfied:

\setlength{\arraycolsep}{0.0em}
\begin{equation}
	\text{dist}(c_t,c_d) \leq \text{dist}_{\max}.
\end{equation}

\noindent In our research, we set \( \text{dist}_{\max} = 0.04 \) meters.
Note that multiple 3D detections can be assigned to a single track. This could be prevented using the Hungarian algorithm, as commonly adopted in many MOT approaches \cite{mot8, mot10}. However, we use a constant threshold to improve the interpretability of associations in environments with densely clustered cherry tomatoes and frequent occlusions. Given a matched pair of a track and a 3D detection, the position and the volume of \( c_k \) are updated as follows:

\setlength{\arraycolsep}{0.0em}
\begin{equation}
	p_{c_t} := (1 - w_p) p_{c_t} + w_p p_{c_d},
\end{equation}
\setlength{\arraycolsep}{0.0em}
\begin{equation} \label{eq:update}
	v_{c_t} := (1 - w_v) v_{c_t} + w_v v_{c_d},
\end{equation}
\noindent where \( w_p \) and \( w_v \) denote the update weights, both set to 0.7. If a detection is not matched with any track, it is initialized as a new track. If a track remains unmatched, it is not updated. Finally, if the number of associated detections in a track exceeds a certain threshold, the track is considered sufficiently reliable, and its 3D context is included in the final output. The required number of associations depends on the duration of the target's visibility. Through empirical tuning, this value is set to 3 in our setting.

\subsection{Yield Estimation}  
\begin{figure*}[tb] 
	\centering
	\includegraphics[width=1.0\linewidth]{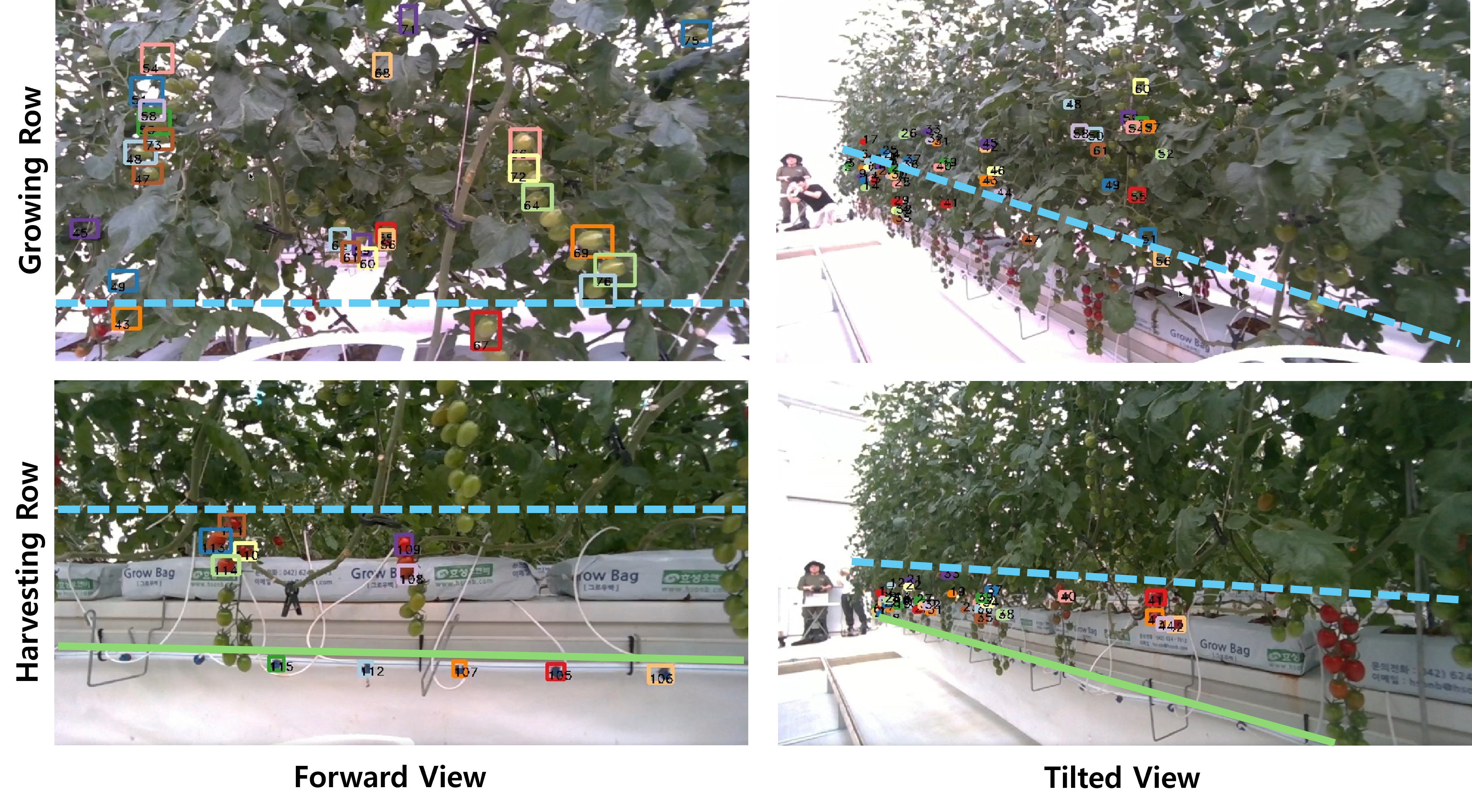}
	\vspace{-6mm}
	\caption{Qualitative results on tracking ripened and unripened cherry tomatoes. All visible tracks are reprojected onto the images with their corresponding IDs. A total of four types of dataset were collected. The blue line represents the boundary between the growing row and the harvesting row, as shown in \figref{fig:farm_analysis}. The light green line indicates the minimum height threshold used to reject false positive tracks (e.g., IDs 105, 106, 107, 112, and 115 in the bottom-left case) during experiments in the harvesting row.
	}
	\vspace{-4mm}
	\label{fig:qualitative}
	\vspace{0mm}
\end{figure*}

The outputs of 3D multi-object tracking are further processed for yield estimation. To estimate the yield, we compute the number and total weight of cherry tomatoes. Since the tracks are obtained through offline 3D MOT processing, those outside the plants are excluded using predefined coordinate constraints, as illustrated in \figref{fig:qualitative}. Additionally, since ripened cherry tomatoes exceed a certain size, we discard tracks with a volume smaller than \(12 \, \text{cm}^3\). The remaining tracks are then regarded as harvesting targets, and their count is used to estimate the number of cherry tomatoes.

{\col 
Since the resulting 3D context consists only of position and volume, the weight of cherry tomatoes cannot be directly obtained. Therefore, the height updated by Eq.~\ref{eq:update} is mapped to weight.  
To find the mapping, three random harvested cherry tomatoes were selected, with heights of 35~mm, 40~mm, and 42~mm, and corresponding weights of 13.5~g, 18.1~g, and 23~g, respectively. Approximating these measurements with a cubic polynomial gives:
\begin{equation}
	\text{weight} = 0.00178 h_{c_t}^3 + 0.00993 h_{c_t}^2 - 7.36 h_{c_t} + 192,
\end{equation}
\noindent where \( h_{c_t} \) represents the track height in millimeters, and weight is given in grams.  
Through this process, the output of the proposed 3D MOT is converted into data suitable for yield estimation.
}

\section{Experiments}
\subsection{Experimental Setup}
\begin{figure}[tb] 
	\centering
	\includegraphics[width=1.0\linewidth]{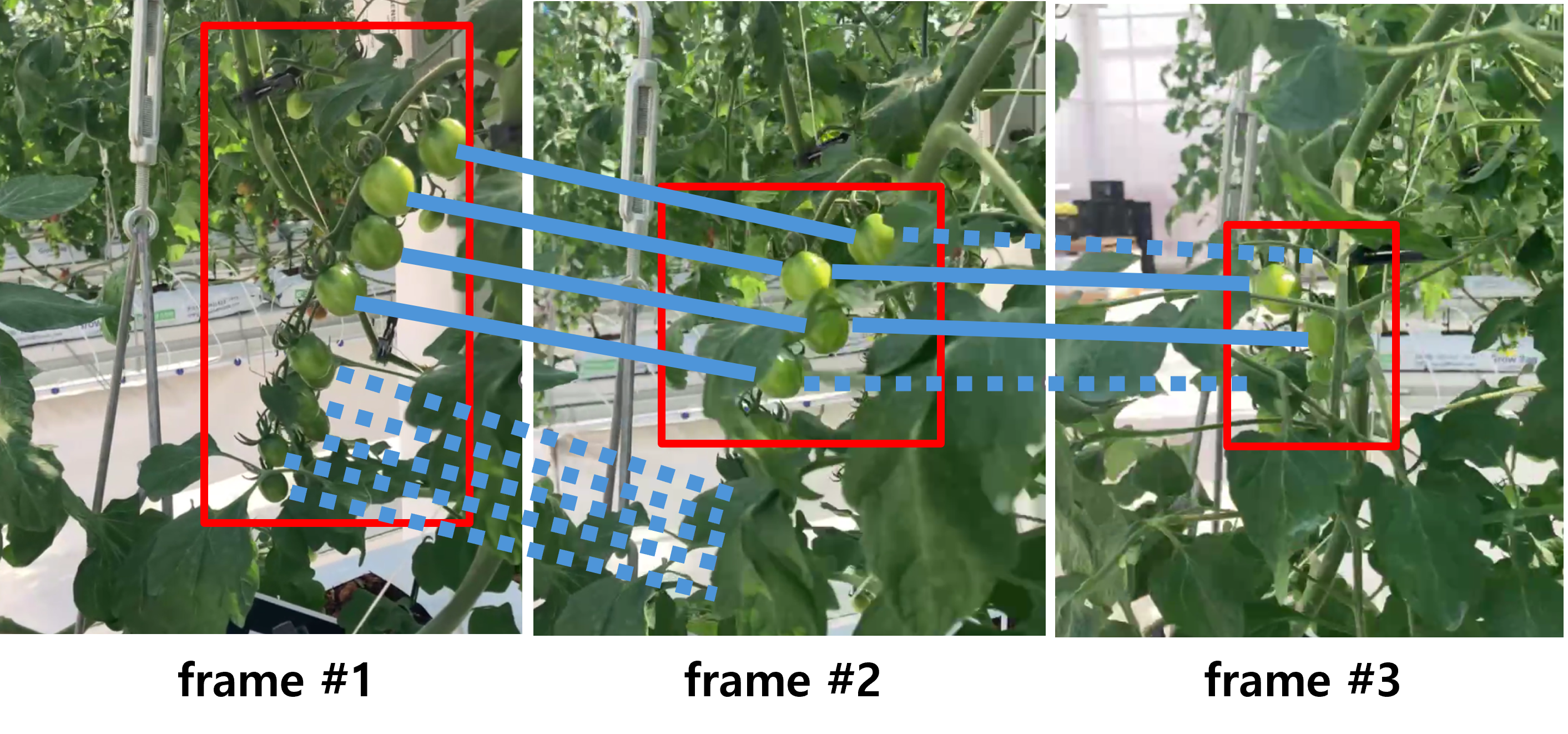}
	\vspace{-6mm}
	\caption{Images of unripened cherry tomatoes from multiple viewpoints. Three frames from a video are presented. Matched fruits from the previous frame are linked with solid blue lines, while unmatched fruits are connected with dotted lines to their expected positions. The observability of the fruits is significantly affected by camera viewpoints.}
	\vspace{-5mm}
	\label{fig:necessity}
\end{figure}

To evaluate the efficiency and effectiveness of UAV-based farm monitoring, we control the UAV manually and collected two dataset for different purposes. In the first experiment, data were captured by the UAV next to the harvesting row to assess the efficiency and accuracy of UAV-based yield estimation. In the second experiment, sensor streams from the growing row were collected, and the applications of 3D MOT were evaluated.

Various viewpoints need be considered to overcome occlusions near target plants in the farm environment for effective monitoring. As shown in \figref{fig:necessity}, the observability of each tomato is significantly affected by the camera viewpoint. To address this, each dataset is collected using two different camera settings: forward view and tilted view, as shown in \figref{fig:qualitative}. In the first experiment, the efficiency and accuracy of yield estimation using the proposed system are evaluated. In the second experiment, insights gained from tracking unripened cherry tomatoes are presented. However, due to challenges in validation, results only from the forward-view camera are validated.

\begin{table*}[t]
	\centering
	\renewcommand{\arraystretch}{1.2}
	\setlength{\tabcolsep}{12pt}
	\caption{Comparison of flight configurations and performance metrics}
	\begin{tabular}{c c c c c}
		\hline
		\textbf{Type} & \textbf{Lane Length (m)} & \textbf{Flight Time (s)} & \textbf{Estimated Count / Error} & \textbf{Avg. Weight Estimation (g) / Error} \\ 
		\hline
		True    & \multirow{3}{*}{13.2} & -      & 89          & 19.14 \\ 
		Forward &                       & 10.449 & 94 / 5.6\%  & 21.53 / 12.5\% \\ 
		Tilted  &                       & 7.149  & 69 / 22.4\% & 26.05 / 36.1\% \\ 
		\hline
	\end{tabular}
	\vspace{-2mm}
	\label{tab:flight_comparison}
\end{table*}

\subsection{Efficiency and Accuracy of UAV-based Yield Estimation} \label{res:yield}
In the first experiment, the efficiency and accuracy of UAV-based yield estimation are validated. The dataset used for evaluation was collected by the UAV flying next to the harvesting row. All ripened cherry tomatoes were harvested for validation, with a total weight of 1704.4 grams and a count of 89. To collect the dataset, the UAV was manually controlled in position mode using LiDAR-inertial-odometry-based pose feedback.

The flight time, estimated count, and predicted average weight are presented in Table \ref{tab:flight_comparison}. Note that the average weight estimation, rather than the total weight sum, is shown to account for inaccuracies in fruit counting. 
The results indicate that our system effectively predicts yield with high accuracy. The required flight time to monitor a 13.2 meter lane is only 10.5 seconds and 7.149 seconds in each experiment. 

When it comes to counting performance, the accuracy in each experiment is 94.4\% and 77.6\%, respectively. In the experiment with the forward-view camera, the estimated count exceeds the ground truth for two reasons. First, a few unripened fruits are misclassified as cherry tomatoes due to repeated false alarms from the YOLOv10 detector. Second, multiple tracks are sometimes generated for a single fruit. This \textit{so-called} double-counting issue occurs in the 3D object detection step.  
When a tomato is briefly occluded by other fruits or thin leaves along the camera's line of sight, its median depth may be incorrectly estimated, leading to an inaccurate 3D detection position. If the 3D detection cube is sufficiently close to a previously initialized track, then both are matched, causing the track's position to be updated incorrectly. After several updates, the misplacement accumulates, potentially leading to the generation of a new track for a fruit that has already been tracked.  
To address this limitation, multiple representative depth values can be used to generate multiple 3D detections. Applying a clustering algorithm such as DBSCAN \cite{dbscan} to the depth image ROI may contribute to this sampling process. 
In the experiment with the tilted-view camera, two challenges in the dataset contributed to lower performance. First, the UAV traveled at a higher velocity, {\col due to inconsistent control during manual flight}. Since our tracking approach relies on repeated observations for association, the number of opportunities to match a 3D detection with a track decreases, leading to a higher number of unreliable tracks. Second, the maximum depth range of our camera is limited, resulting in failure of 3D detection. Even if fruits are localized in the color image, they may be rejected due to invalid depth values, which does not occur in the forward-view experiment. Note that we set the maximum accepted depth to 1 m, exceeding the D405 specification, as based on iterative tuning with the dataset.

For weight estimation, the estimated weight exceeded the ground truth by 12.5\% and 36.1\% in each experiment. We suspect that this discrepancy originates from inaccuracies in mapping cherry tomato height to weight. If a more accurate mapping function is derived, the error rate is expected to decrease.

\begin{table}[t]
	\centering
	\caption{Tracking results for unripened cherry tomatoes}
	\label{tab:tracking_results}
	\begin{tabular}{c|c|c}
		\hline
		\textbf{Timestamp (s)} & \textbf{\makecell{Unripened Tomatoes \\ in Left Half Plane}} & \textbf{\# of Positive Tracks} \\
		\hline
		26.64 & 12  & 4  \\
		27.64 & 34  & 16 \\
		28.64 & 7   & 3  \\
		29.64 & 11  & 5  \\
		30.64 & 20  & 8  \\
		31.64 & 24  & 13 \\
		32.64 & 17  & 6  \\
		33.64 & 31  & 10 \\
		34.64 & 13  & 3  \\
		35.64 & 24  & 14 \\
		\hline
		\textbf{Total} & \textbf{193} & \textbf{82} \\
		\hline 
	\end{tabular}
	\vspace{-8mm}
\end{table}

\subsection{Evaluation of Tracking Densely Located Unripened Cherry Tomato.}
In this section, the accuracy of the proposed multi-object tracking method is evaluated. Manually annotating unripened cherry tomatoes is highly challenging due to their visual similarity to leaves, frequent occlusions, and the limited resolution of depth sensors. These factors make it difficult to obtain precise ground truth data, leading to a simplified evaluation approach. Instead of frame-by-frame annotation, we sampled 10 frames at 1-second interval and verified the accuracy of reprojected 3D tracks. This method allows for practical validation while minimizing annotation complexity in dense farm environments. The results are shown in Table~\ref{tab:tracking_results}.

The resulting counting performance is 42.5\%. This unsatisfactory results is affected by several factors. First, unripened cherry tomatoes are difficult to distinguish from surrounding leaves, and  poor lighting due to lush leaves further reduces contrast, making detection less reliable. Second, frequent occlusions limit the number of effective observations, making data association more difficult and increasing the chances of missing or inconsistently tracking fruits. Unlike controlled environments, real-world farm settings introduce unpredictable lighting, dense plant structures, and sensor limitations, all of which contribute to tracking errors. Additionally, inaccuracies in depth estimation can cause positional errors in 3D space, further degrading tracking performance. These challenges highlight the need for improved viewpoint selection, adaptive perception strategies, and robust multi-view integration to enhance tracking accuracy in complex agricultural environments.

\section{Conclusions}
To address the time inefficiency of UGV-based greenhouse monitoring, we designed a UAV-based indoor farm monitoring system. The proposed UAV is equipped with a camera, a 3D LiDAR, and an IMU. It weighs only 1.4 kg, has a 0.45 m wheelbase, and can fly over 1 km. A 3D MOT algorithm is introduced to identify and localize the target’s 3D position using \mbox{RGB-D} images and the UAV's 6D pose. Additionally, a simple mapping between tomato size and weight is proposed. To validate the system, the UAV monitored a harvesting lane in a real cherry tomato farm. During a 13.2 m flight completed in just 10.5 seconds, the system achieved a counting accuracy of 94.4\% and a weight estimation accuracy of 87.5\%. The proposed system was further applied to the tracking of unripened cherry tomatoes, a particularly challenging task due to high occlusion and visual ambiguity. Through this experiment, we identified the primary sources of tracking difficulties and derived key requirements for active perception in greenhouse environments.

To the best of our knowledge, this is among the earlier studies to address the efficiency of UAV-based, GNSS-denied indoor farm monitoring. However, due to the extensive labor required for manual annotation, our validation was conducted only partially. Automating the generation of validation dataset is essential to further enhance the efficiency of agricultural monitoring.

UAV-based greenhouse inspection presents significant challenges but remains a highly attractive solution. To further optimize its efficiency in densely occluded farm environments, an effective strategy for selecting camera viewpoints need be considered.

	\bibliographystyle{IEEEtran}

\end{document}